\documentclass[10pt, a4paper]{article}
\usepackage{lrec}
\usepackage{multibib}
\newcites{languageresource}{Language Resources}
\usepackage{graphicx}
\usepackage{tabularx}
\usepackage{soul}
\usepackage{epstopdf}
\usepackage[utf8]{inputenc}

\usepackage{hyperref}
\usepackage{xstring}
\usepackage{svg}

\usepackage{color}

\title{ZuCo 2.0: A Dataset of Physiological Recordings \\During Natural Reading and Annotation}

\name{Nora Hollenstein\textsuperscript{1}, Marius Troendle\textsuperscript{2}, Ce Zhang\textsuperscript{1}, Nicolas Langer\textsuperscript{2}}

\address{\textsuperscript{1} Department of Computer Science, ETH Zurich  \\
         \textsuperscript{2} Department of Psychology, University of Zurich \\
         \{noraho,ce.zhang\}@inf.ethz.ch\\
         \{m.troendle,n.langer\}@psychologie.uzh.ch\\}

\abstract{
We recorded and preprocessed ZuCo 2.0, a new dataset of simultaneous eye-tracking and electroencephalography during natural reading and during annotation. This corpus contains gaze and brain activity data of 739 English sentences, 349 in a normal reading paradigm and 390 in a task-specific paradigm, in which the 18 participants actively search for a semantic relation type in the given sentences as a linguistic annotation task. This new dataset complements ZuCo 1.0 by providing experiments designed to analyze the differences in cognitive processing between natural reading and annotation. The data is freely available here: \url{https://osf.io/2urht/}. \\ 
\newline  }

\begin{document}

\maketitleabstract

\section{Introduction}

How humans process language has become increasingly relevant in natural language processing (NLP) since physiological data during language understanding is more accessible and recorded with less effort. In this work, we focus on eye-tracking and electroencephalography (EEG) recordings to capture the reading process. On one hand, eye movement data provides millisecond-accurate records about where humans look when they are reading, and is highly correlated with the cognitive load associated with different stages of text processing. On the other hand, EEG records electrical brain activity across the scalp and is a direct measure of physiological processes, including language processing. The combination of both measurement methods enables us to study the language understanding process in a more natural setting, where participants read full sentences at a time, in their own speed. Eye-tracking then permits us to define exact word boundaries in the timeline of a subject reading a sentence, allowing the extraction of brain activity signals for each word.

\begin{figure}[t]
    \centering
    \includegraphics[width=0.367\textwidth]{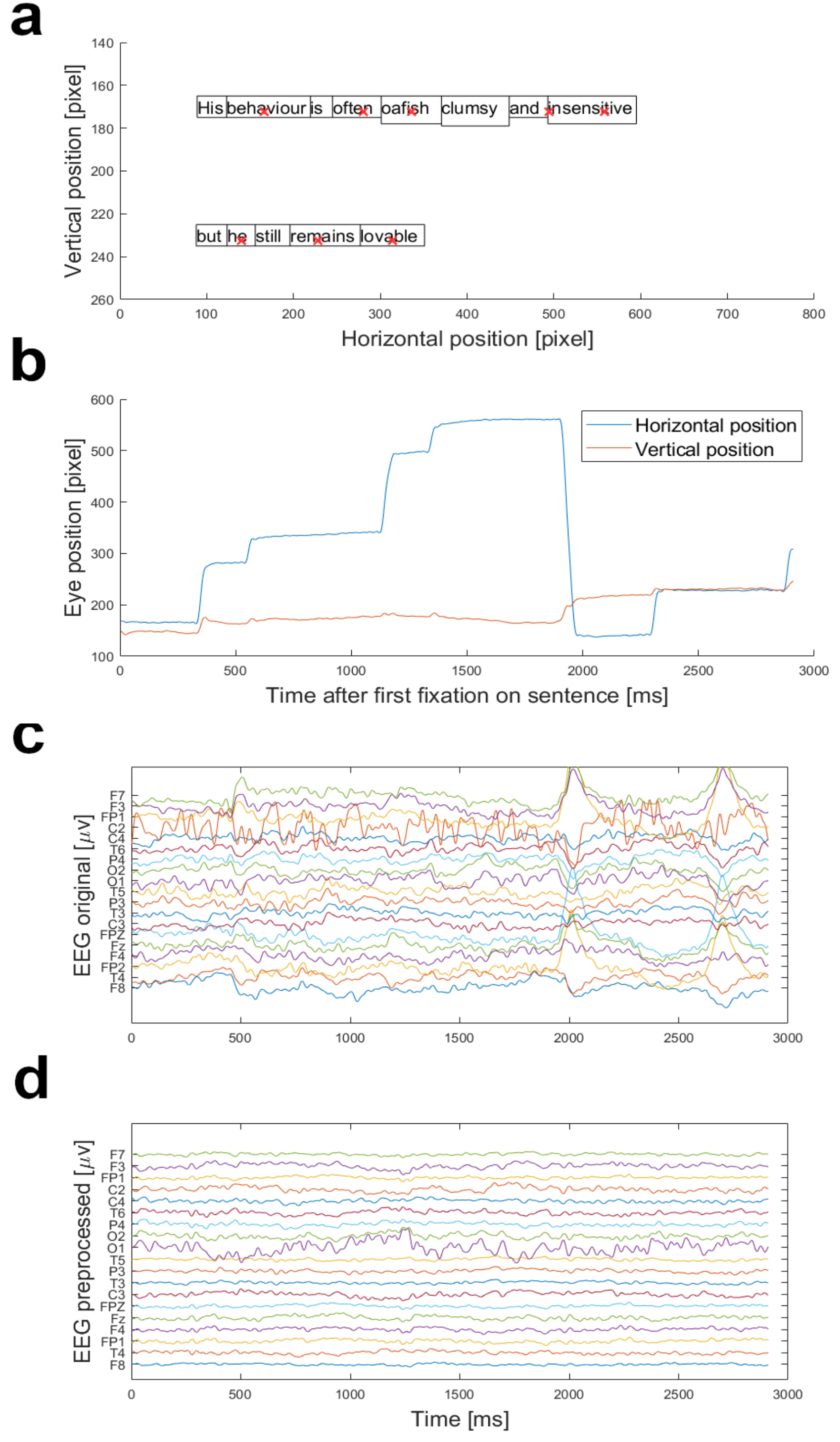} 
    \caption{Visualization of eye-tracking and EEG data for a single sentence. (a) Prototypical sentence fixation data. Red crosses indicate fixations; boxes around the words indicate the wordbounds. (b) Fixation data plotted over time. (c) Raw EEG data during a single sentence. (d) Same data as in (c) after preprocessing.}
    \label{fig:example}
\end{figure}

Human cognitive language processing data is immensely useful for NLP: Not only can it be leveraged to improve NLP applications (e.g. \newcite{barrett2016weakly} for part-of-speech tagging, or \newcite{klerke2016improving} for sentence compression), but also to evaluate state-of-the-art machine learning systems. For example, \newcite{hollenstein2019cognival} evaluate word embeddings, or \newcite{schwartz2019inducing} fine-tune language models with brain-relevant bias.

Additionally, the availability of labelled data plays a crucial role in all supervised machine learning applications. Physiological data can be used to understand and improve the labelling process (e.g. \newcite{tokunaga2017eye}), and, for instance, to build cost models for active learning scenarios \cite{tomanek2010cognitive}. Is it possible to replace this expensive manual work with models trained on physiological activity data recorded from humans while reading? That is to say, can we find and extract the relevant aspects of text understanding and annotation directly from the source, i.e. eye-tracking and brain activity signals during reading?

\begin{table*}[t]
\centering
\begin{tabular}{|lcllcccc|}
\hline
\textbf{ID} & \textbf{Age} & \textbf{Gender} & \textbf{LexTALE} & \textbf{Score NR} & \textbf{Score TSR} & \textbf{Speed NR} & \textbf{Speed TSR} \\\hline\hline
YAC & 32 & female & 76.25\% & 82.61\% & 83.85\% & 5.27 & 4.96 \\
YAG & 47 & female & 93.75\% & 91.30\% & 56.92\% & 7.64 & 8.73 \\
YAK & 31 & female & 100.00\% & 74.07\% & 96.41\% & 3.83 & 5.89 \\
YDG & 51 & male & 100.00\% & 91.30\% & 96.67\% & 4.97 & 3.93 \\
YDR & 25 & male & 85.00\% & 78.26\% & 96.92\% & 4.32 & 2.32 \\
YFR & 27 & male & 85.00\% & 89.13\% & 94.36\% & 6.48 & 4.79 \\
YFS & 39 & male & 90.00\% & 91.30\% & 96.15\% & 3.96 & 2.85 \\
YHS & 31 & male & 90.00\% & 78.26\% & 97.69\% & 3.30 & 2.40 \\
YIS & 52 & male & 97.50\% & 89.13\% & 98.46\% & 5.82 & 2.58 \\
YLS & 34 & female & 93.75\% & 91.30\% & 92.31\% & 5.57 & 5.85 \\
YMD & 31 & female & 100.00\% & 86.96\% & 95.64\% & 7.50 & 6.24 \\
YMS & 36 & female & 86.25\% & 89.13\% & 95.38\% & 7.68 & 3.35 \\
YRH & 28 & female & 81.25\% & 86.96\% & 95.64\% & 5.14 & 4.32 \\
YRK & 29 & female & 85.00\% & 97.83\% & 96.15\% & 7.35 & 7.70 \\
YRP & 23 & female & 82.50\% & 78.26\% & 90.00\% & 7.14 & 8.37 \\
YSD & 34 & male & 95.00\% & 93.48\% & 94.36\% & 5.01 & 2.87 \\
YSL & 32 & female & 71.25\% & 84.78\% & 83.85\% & 6.73 & 6.14 \\
YTL* & 36 & male & 81.25\% & 80.43\% & 94.10\% & 7.48 & 3.23 \\\hline
\textbf{mean} & \textbf{34} & \textbf{44\% male} & \textbf{88.54\%} & \textbf{86.36\%} & \textbf{91.94\%} & \textbf{5.84} & \textbf{4.81} \\\hline
\end{tabular}
\caption{Subject demographics, LexTALE scores, and control scores and reading speed (i.e. seconds per sentence) for each task. The * next to the subject ID marks a bilingual subject.}
\label{tab:subjects}
\end{table*}

Motivated by these questions and our previously released dataset, ZuCo 1.0 \cite{hollenstein2018zuco}, we developed this new corpus, where we specifically aim to collect recordings during natural reading as well as during annotation. \\

We provide the first dataset of simultaneous eye movement and brain activity recordings to analyze and compare normal reading to task-specific reading during annotation. The \textbf{Zurich Cognitive Language Processing Corpus (ZuCo) 2.0}, including raw and preprocessed eye-tracking and electroencephalography (EEG) data of 18 subjects, as well as the recording and preprocessing scripts, is publicly available at \url{https://osf.io/2urht/}. It contains physiological data of each subject reading 739 English sentences from Wikipedia (see example in Figure \ref{fig:example}). We want to highlight the re-use potential of this data. In addition to the neuro- and psycho-linguistic motivation, this corpus is especially tailored for training and evaluating machine learning algorithms for NLP purposes. It allows to conduct experiments for different semantic tasks, such as information extraction, including entity and relation discovery.\\

In this work, we describe the corpus construction, including experimental design and data acquisition. Further, we explain which preprocessing and feature extraction steps were applied. And finally, we conduct a detailed technical validation of the data as proof of the quality of the recordings.

\section{Related Work}
Some eye-tracking corpora of natural reading
(e.g. the Dundee corpus \cite{kennedy2003dundee}, Provo corpus \cite{luke2017provo} and GECO corpus \cite{cop2017presenting}), and a few EEG corpora (for example, the UCL corpus by \newcite{frank2017wordpred}) are available. It has been shown that this type of cognitive processing data is useful for improving and evaluating NLP methods (e.g. \newcite{long2017cognition}, \newcite{barrett2018sequence}, \newcite{hale2018finding}, \newcite{hollenstein2019cognival}). However, before the Zurich Cognitive Language Processing Corpus (ZuCo 1.0), there was no available data for simultaneous eye-tracking and EEG recordings of natural reading. \newcite{dimigen2011coregistration} studied the linguistic effects of eye movements and EEG co-registration in natural reading and showed that they accurately represent lexical processing. Moreover, the simultaneous recordings are crucial to extract word-level brain activity signals.

While the above mentioned studies analyze and leverage natural reading, some NLP work has used eye-tracking during annotation (but, as of yet, not EEG data). \newcite{mishra2016predicting} and \newcite{joshi2014measuring} recorded eye-tracking during binary sentiment annotation (positive/negative). This data was used to determine the annotation complexity of the text passages based on eye movement metrics \cite{mishra2017scanpath} and for sarcasm detection \cite{mishra2017harnessing}. Moreover, eye-tracking has been used to analyze the word sense annotation process in Hindi \cite{joshi2013more}, named entity annotation in Japanese \cite{tokunaga2017eye}, and to leverage annotator gaze behaviour for English coreference resolution \cite{cheri2016leveraging}. Finally, \newcite{tomanek2010cognitive} used eye-tracking data during entity annotation to build a cost model for active learning. However, until now there was no available data or research that analyzes the differences in the human processing of normal reading versus annotation.\\

\begin{table}[t]
\centering
\begin{tabular}{|l|l|l|}
\hline
& \textbf{NR} & \textbf{TSR} \\\hline \hline
sentences & 349   & 390   \\\hline
sent. length & mean (SD), range & mean (SD), range \\
 & 19.6 (8.8), 5-53 & 21.3 (9.5), 5-53 \\\hline
total words & 6828 & 8310 \\
word types & 2412 & 2437 \\\hline
word length & mean (SD), range & mean (SD), range \\
 & 4.9 (2.7), 1-29 & 4.9 (2.7), 1-21 \\\hline
 Flesch score & 55.38 &  50.76  \\
 \hline
\end{tabular}
\caption{Descriptive statistics of reading materials (SD = standard deviation), including Flesch readibility scores.}
\label{tab:datastats}
\end{table}

\begin{table}[t]
\centering
\begin{tabular}{|l|c|}
\hline
\textbf{Relation type} & \textbf{Sentences} \\\hline\hline
Political affiliation & 45 (9) \\
Education & 72 (10) \\
Wife & 54 (12) \\
Job title & 65 (11) \\
Employer & 54 (10) \\
Nationality & 60 (8) \\
Founder & 40 (8) \\\hline
\textbf{total} & \textbf{390 (68)} \\\hline
\end{tabular}
\caption{Distribution of relation types in the task-specific reading. The right column contains the number of sentences, and the number control sentences without a relation in brackets.}
\label{tab:relations}
\end{table}

\paragraph{ZuCo1.0}
In previous work, we recorded a first dataset of simultaneous eye-tracking and EEG during natural reading \cite{hollenstein2018zuco}. ZuCo 1.0\footnote{Data available here: \url{https://osf.io/q3zws/}} consists of three reading tasks, two of which contain very similar reading material and experiments as presented in the current work. However, the main difference and reason for recording ZuCo 2.0 consists in the experiment procedure. For ZuCo 1.0 the normal reading and task-specific reading paradigms were recorded in different sessions on different days. Therefore, the recorded data is not appropriate as a means of comparison between natural reading and annotation, since the differences in the brain activity data might result mostly from the different sessions due to the sensitivity of EEG. This, and extending the dataset with more sentences and more subjects, were the main factors for recording the current corpus. We purposefully maintained an overlap of some sentences between both datasets to allow additional analyses (details are described in Section \ref{sec:materials}).

\section{Corpus Construction}

In this section we describe the contents and experimental design of the ZuCo 2.0 corpus.

\subsection{Participants}
We recorded data from 19 participants and discarded the data of one of them due to technical difficulties with the eye-tracking calibration. Hence, we share the data of 18 participants. All participants are healthy adults (mean age = 34 (SD=8.3), 10 females). Their native language is English, originating from Australia, Canada, UK, USA or South Africa. Two participants are left-handed and three participants wear glasses for reading. Details on subject demographics can be found in Table \ref{tab:subjects}. All participants gave written consent for their participation and the re-use of the data prior to the start of the experiments. The study was approved by the Ethics Commission of the University of Zurich.\\
%Furthermore, the participants also reported their highest level of education, i.e. the highest degree achieved. 10 participants completed Bachelor studies, 5 Master studies and 2 PhD.\\ %correlation between this and lextale score? reading speed? control answers?

\subsection{Reading materials}\label{sec:materials}

\begin{figure*}[t]
    \centering
    \includegraphics[width=1\textwidth]{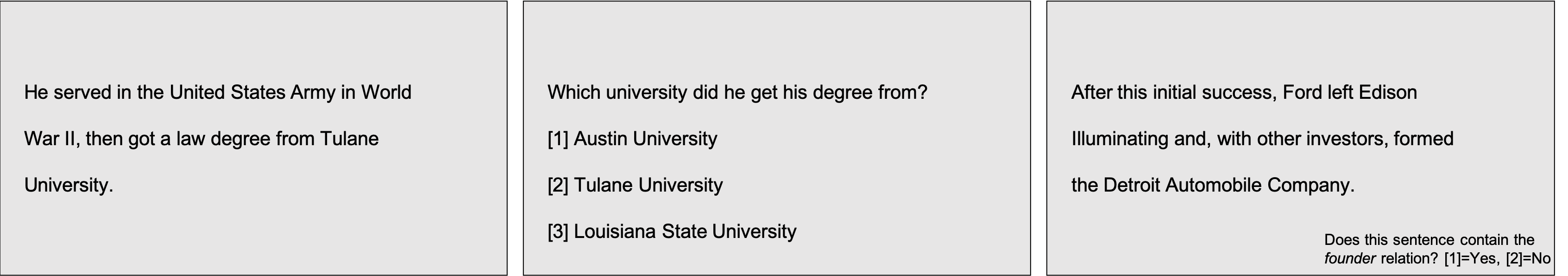} 
    \caption{Example sentences on the recording screen: (left) a normal reading sentence, (middle) a control question for a normal reading sentence, and (right) a task-specific  annotation sentence.}
    \label{fig:sentences}
\end{figure*}

During the recording session, the participants read 739 sentences that were selected from the Wikipedia corpus provided by \newcite{culotta2006integrating}. This corpus was chosen because it provides annotations of semantic relations. Relation detection is a high-level semantic task requiring complex cognitive processing. We included seven of the originally defined relation types: \textit{political\_affiliation}, \textit{education}, \textit{founder}, \textit{wife/husband}, \textit{job\_title}, \textit{nationality}, and \textit{employer}. The sentences were chosen in the same length range as ZuCo 1.0, and with similar Flesch reading ease scores \cite{kincaid1975derivation}. The dataset statistics are shown in Table \ref{tab:datastats}.

Of the 739 sentences, the participants read 349 sentences in a normal reading paradigm, and 390 sentences in a task-specific reading paradigm, in which they had to determine whether a certain relation type occurred in the sentence or not. Table \ref{tab:relations} shows the distribution of the different relation types in the sentences of the task-specific annotation paradigm.

Purposefully, there are 63 duplicates between the normal reading and the task-specific sentences (8\% of all sentences). The intention of these duplicate sentences is to provide a set of sentences read twice by all participants with a different task in mind. Hence, this enables the comparison of eye-tracking and brain activity data when reading normally and when annotating specific relations (see examples in Section \ref{validation}).

Furthermore, there is also an overlap in the sentences between ZuCo 1.0 and ZuCo 2.0. 100 normal reading and 85 task-specific sentences recorded for this dataset were already recorded in ZuCo 1.0. This allows for comparisons between the different recording procedures (i.e. session-specific effects) and between more participants (subject-specific effects).

\subsection{Experimental design}
%task overview, stimuli and experimental procedure, participant instructions \\
%materials: example sentence for each task, example of control questions, \\
%Figure \ref{fig:sentences} shows examples sentences for both tasks.

As mentioned above, we recorded two different reading tasks for the ZuCo 2.0 dataset. During both tasks the participants were able to read in their own speed, using a control pad to move to the next sentence and to answer the control questions, which allowed for natural reading. Since each subject reads at their own personal pace, the reading speed between varies between subjects. Table \ref{tab:subjects} shows the average reading speed for each task, i.e. the average number of seconds a subject spends per sentence before switching to the next one.

\begin{figure*}[t]
    \centering
   \includegraphics[width=0.33\textwidth]{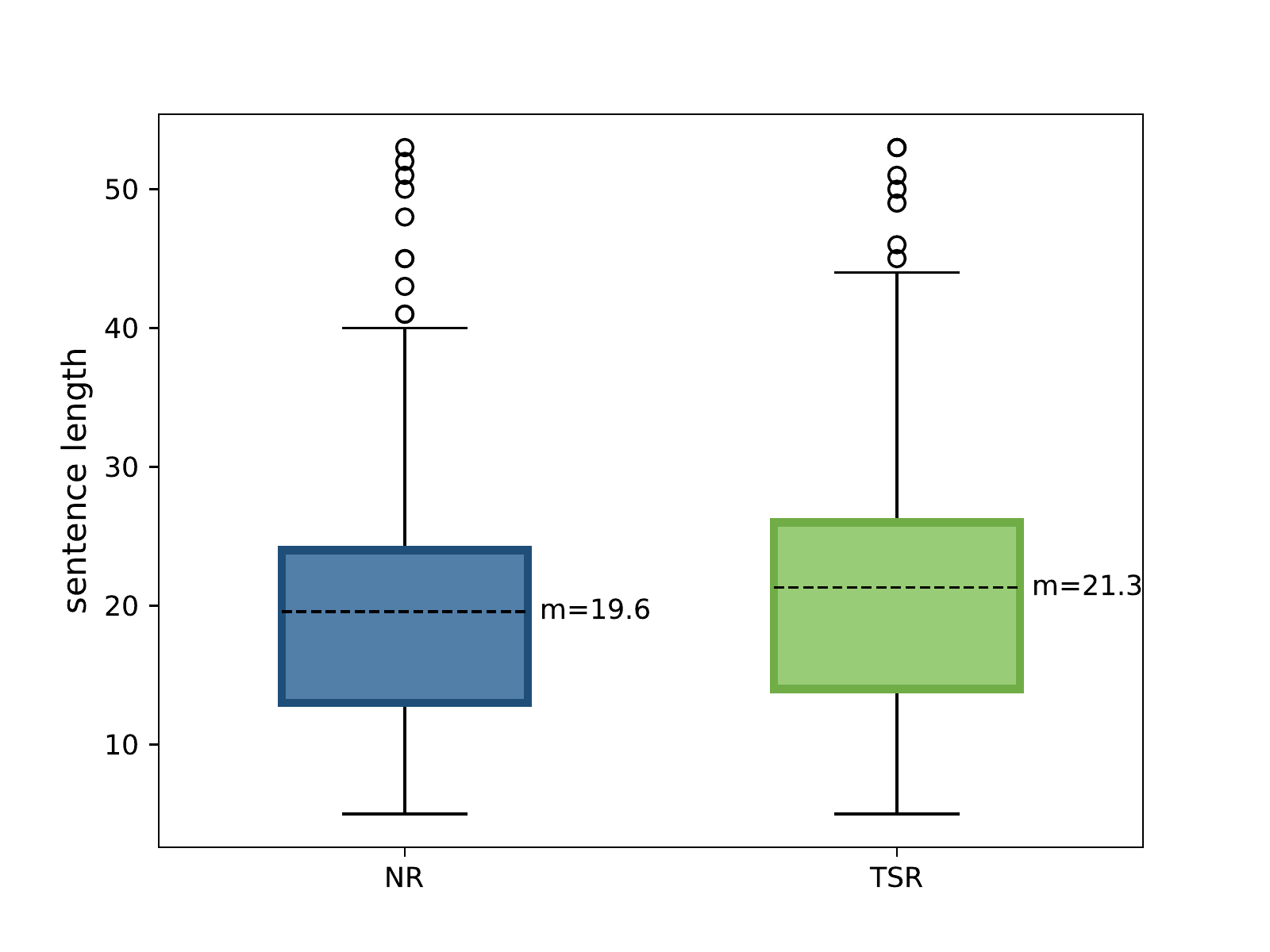}
   \includegraphics[width=0.33\textwidth]{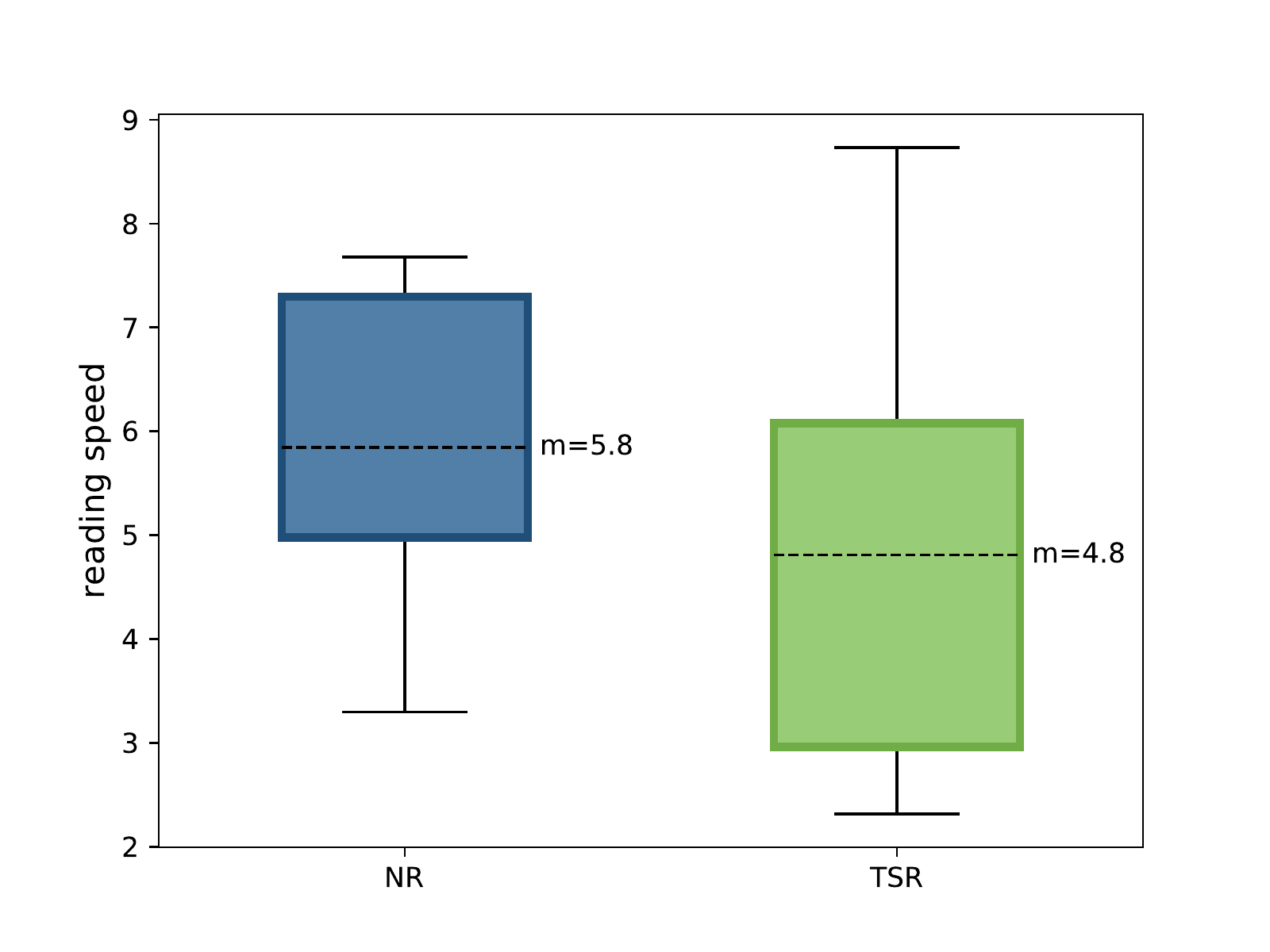}
   \includegraphics[width=0.33\textwidth]{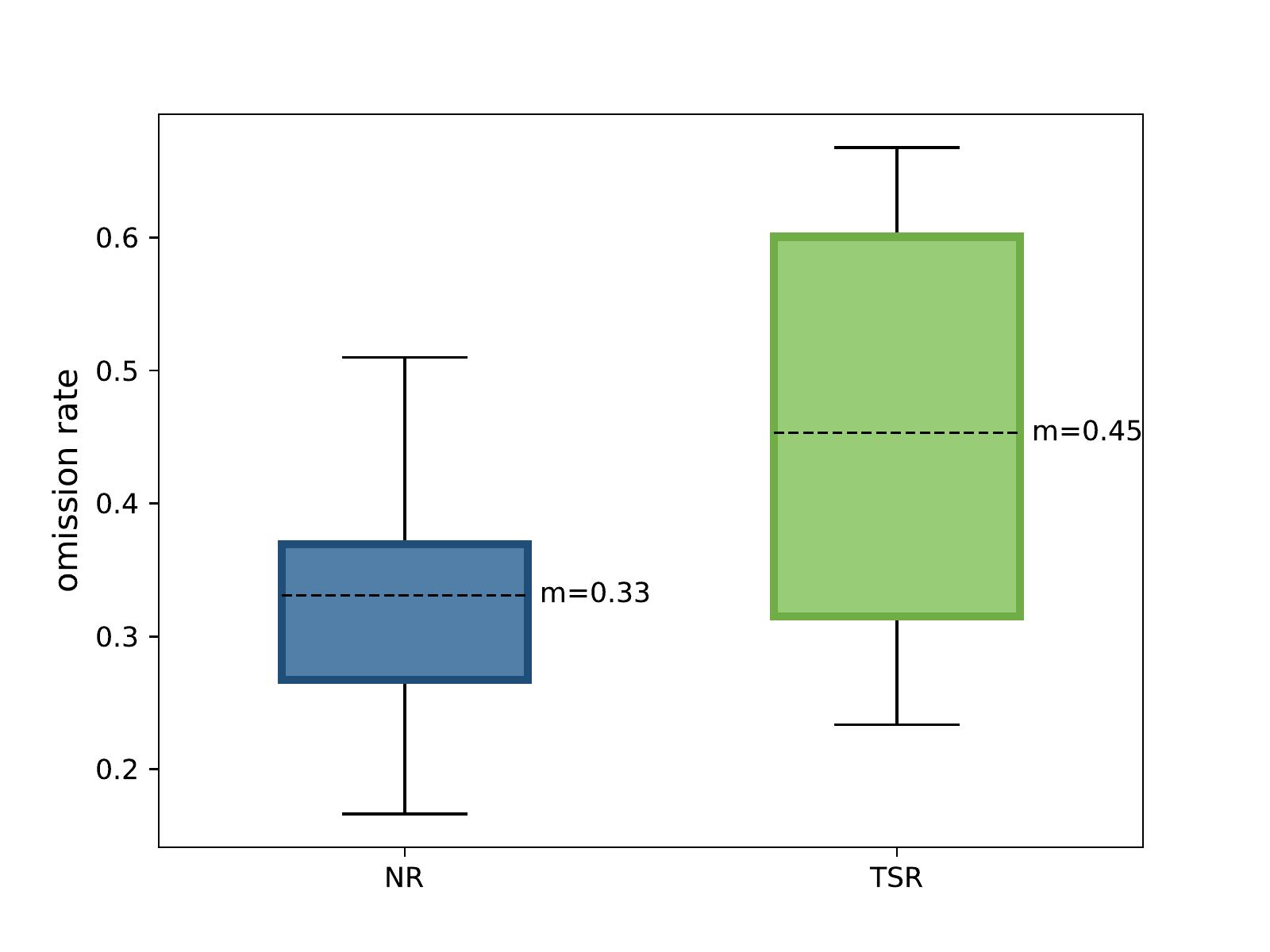}
    \caption{Sentence length (words per sentence), reading speed (seconds per sentence) and omission rate (percentage of words not fixated) comparison between normal reading (NR) and task-specific reading (TSR).}
    \label{fig:omr}
\end{figure*}

All 739 sentences were recorded in a single session for each participant. The duration of the recording sessions was between 100 and 180 minutes, depending on the time required to set up and calibrate the devices, and the personal reading speed of the participants.

We recorded 14 blocks of approx. 50 sentences, alternating between tasks: 50 sentences of normal reading, followed by 50 sentences of task-specific reading. The order of blocks and sentences within blocks was identical for all subjects. Each sentence block was preceded by a practice round of three sentences and followed by a short break to ensure a clear separation between the reading tasks.

\paragraph{Normal reading (NR)}

In the first task, participants were instructed to read the sentences naturally, without any specific task other than comprehension. Participants were told to read the sentences normally without any special instructions. Figure \ref{fig:sentences} (left) shows an example sentence as it was depicted on the screen during recording. As shown in Figure \ref{fig:sentences} (middle), the control condition for this task consisted of single-choice questions about the content of the previous sentence. 12\% of randomly selected sentences were followed by such a comprehension question with three answer options on a separate screen. 

\paragraph{Task-specific reading (TSR)}

In the second task, the participants were instructed to search for a specific relation in each sentence they read. Instead of comprehension questions, the participants had to decide for each sentence whether it contains the relation or not, i.e. they were actively annotating each sentence. Figure \ref{fig:sentences} (right) shows an example screen for this task. 17\% of the sentences did not include the relation type and were used as control conditions. All sentences within one block involved the same relation type. The blocks started with a practice round, which described the relation and was followed by three sample sentences, so that the participants would be familiar with the respective relation type.

\begin{figure}[t]
    \centering
    \includegraphics[width=0.4\textwidth]{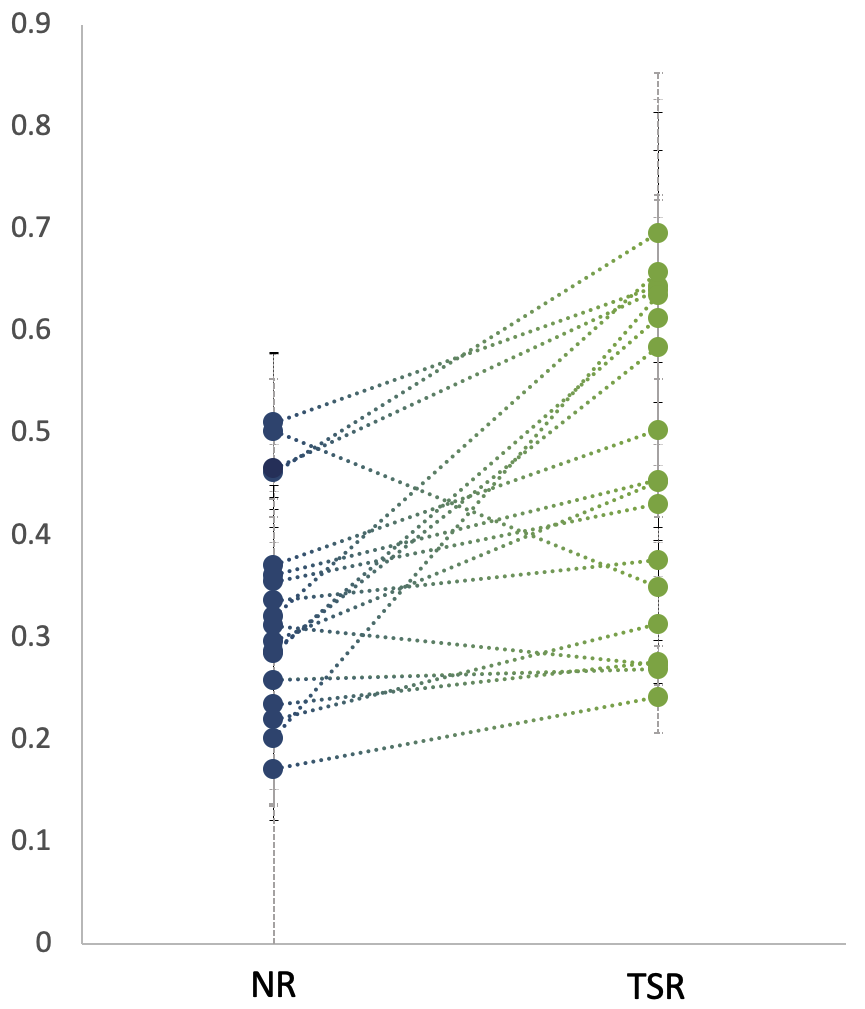}
    \caption{Skipping proportion on word level for both tasks.}
    \label{fig:skipping}
\end{figure}

\subsection{Linguistic assessment}
As a linguistic assessment, the vocabulary and language proficiency of the participants was tested with the LexTALE test (Lexical Test
for Advanced Learners of English, \newcite{lemhofer2012introducing}). This is an unspeeded lexical decision task designed for intermediate to highly proficient language users. The average LexTALE score over all participants was 88.54\%. Moreover, we also report the scores the participants achieved with their answers to the reading comprehension control questions and their relation annotations. The detailed scores for all participants are also presented in Table \ref{tab:subjects}.

\begin{figure*}[t]
    \centering
    \includegraphics[width=1\textwidth]{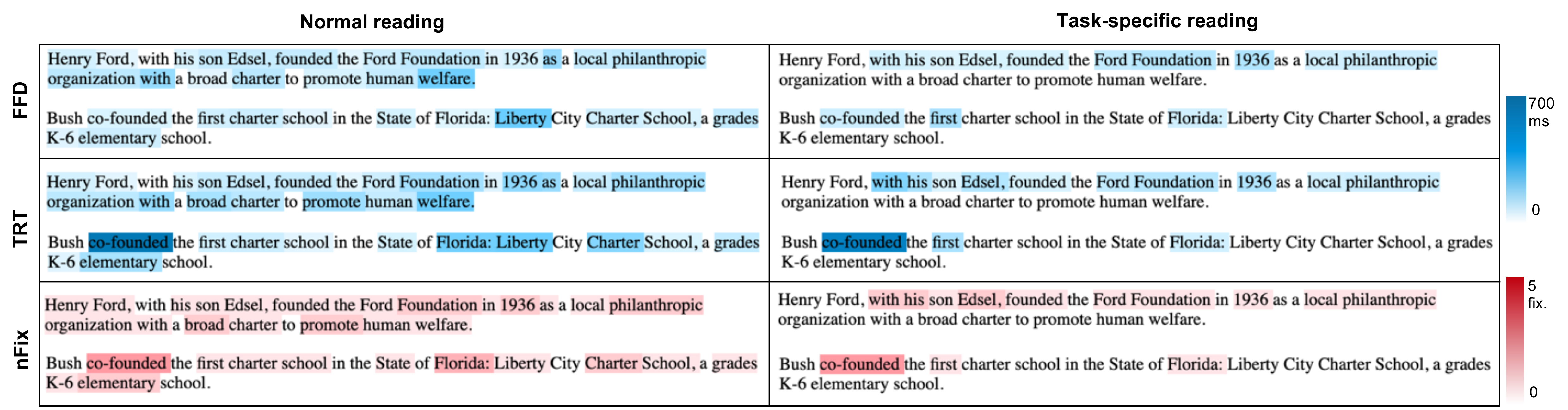} 
    \caption{Fixation heatmaps for two sentences containing the relation \textit{founder}, showing a comparison of the eye-tracking features between normal reading and task-specific annotation reading (first fixation duration (FFD), total reading time (TRT), number of fixations (nFix).}
    \label{fig:heatmaps}
\end{figure*}

\subsection{Data acquisition}

Data acquisition took place in a sound-attenuated and dark experiment room. Participants were seated at a distance of 68cm from a 24-inch monitor with a resolution of 800x600 pixels. A stable head position was ensured via a chin rest. Participants were instructed to stay as still as possible during the tasks to avoid motor EEG artifacts. Participants were also offered snacks and water during the breaks and were encouraged to rest. All sentences were presented at the same position on the screen and could span multiple lines. The sentences were presented in black on a light grey background with font size 20-point Arial, resulting in a letter height of 0.8 mm. The experiment was programmed in MATLAB 2016b \cite{mathworks2000language}, using PsychToolbox \cite{brainard1997psychophysics}. Participants completed the tasks sitting alone in the room, while two research assistants were monitoring their progress in the adjoining room. All recording scripts including detailed participant instructions are available alongside the data.

\paragraph{Eye-tracking acquisition} Eye position and pupil size were recorded with an infrared video-based eye tracker (EyeLink 1000 Plus, SR Research) at a sampling rate of 500 Hz. The eye tracker was calibrated with a 9-point grid at the beginning of the session and re-validated before each block of sentences.

\paragraph{EEG acquisition} High-density EEG data were recorded at a sampling rate of 500 Hz with a bandpass of 0.1 to 100 Hz, using a 128-channel EEG Geodesic Hydrocel system (Electrical Geodesics). The recording reference was set at electrode \textit{Cz}. The head circumference of each participant was measured to select an appropriately sized EEG net. To ensure good contact, the impedance of each electrode was checked prior to recording, and was kept below 40 kOhm. Electrode impedance levels were checked after every third block of 50 sentences (approx. every 30 mins) and reduced if necessary.

\subsection{Preprocessing and feature extraction}

\paragraph{Eye-tracking} The eye-tracking data consists of $(x,y)$ gaze location entries for all individual fixations (Figure \ref{fig:example}b). Coordinates were given in pixels with respect to the monitor coordinates (the upper left corner of the screen was $(0,0)$ and down/right was positive). We provide this raw data as well as various engineered eye-tracking features. For this feature extraction only fixations within the boundaries of each displayed word were extracted. Data points distinctly not associated with reading (minimum distance of 50 pixels to the text) were excluded. Additionally, fixations shorter than 100 ms were excluded from the analyses, because these are unlikely to reflect fixations relevant for reading \cite{sereno2003measuring}.
On the basis of the GECO and ZuCo 1.0 corpora, we extracted the following features: (i) \textit{gaze duration} (GD), the sum of all fixations on the current word in the first-pass reading before the eye moves out of the word; (ii) \textit{total reading time} (TRT), the sum of all fixation durations on the current word, including regressions; (iii) \textit{first fixation duration} (FFD), the duration of the first fixation on the prevailing word; (iv) \textit{single fixation duration} (SFD), the duration of the first and only fixation on the current word; and (v) \textit{go-past time} (GPT), the sum of all fixations prior to progressing to the right of the current word, including regressions to previous words that originated from the current word. For each of these eye-tracking features we additionally computed the pupil size. Furthermore, we extracted the number of fixations and mean pupil size for each word and sentence.

\begin{figure}[t]
    \centering
    \includegraphics[width=0.5\textwidth]{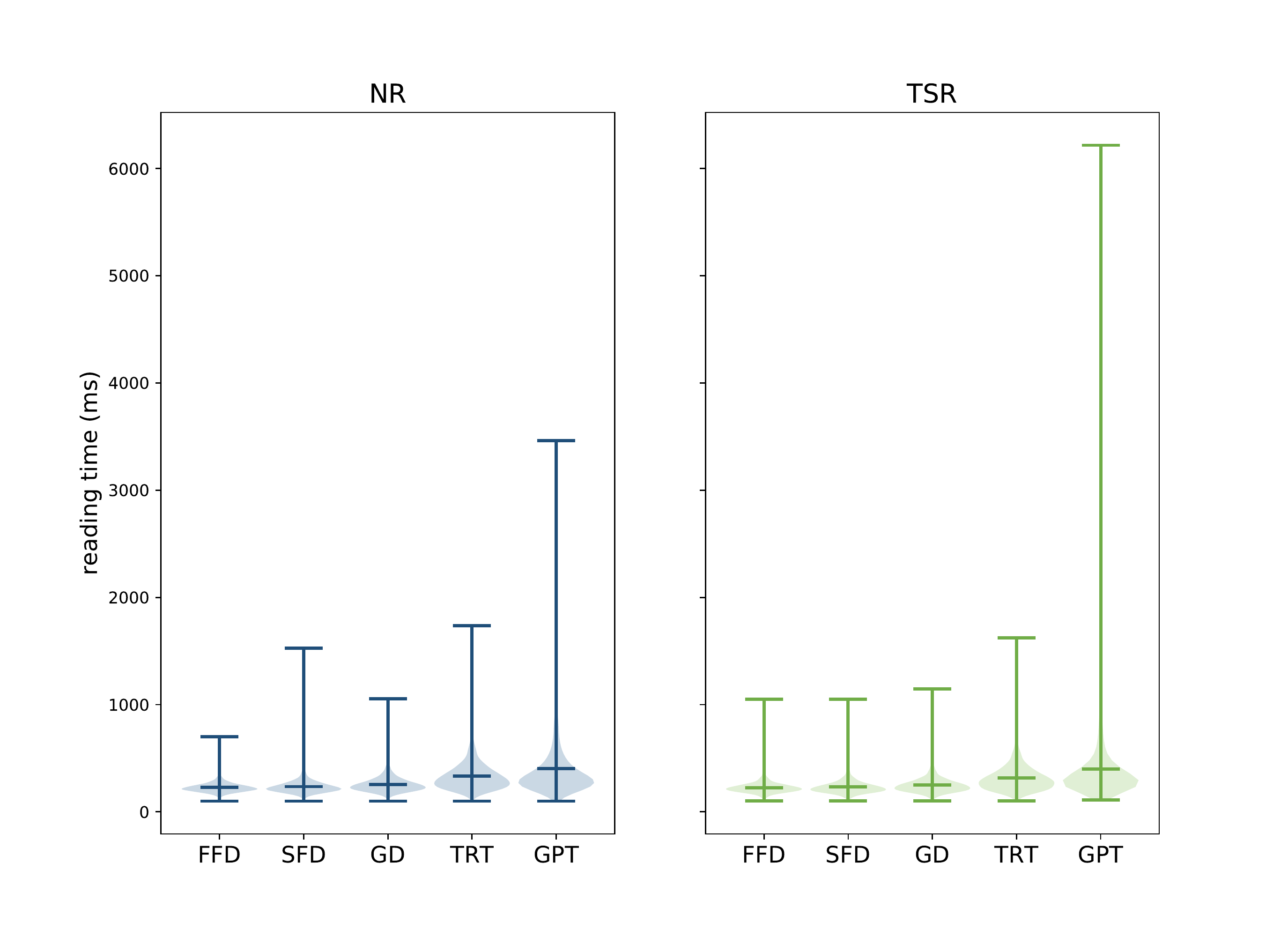} 
    \caption{Violin plots showing means, distributions, and ranges of the reading time measures per word for each task and each eye-tracking feature (x-axis) in milliseconds.}
    \label{fig:violinplots}
\end{figure}

%\begin{figure*}[t]
%    \centering
%    \includegraphics[width=0.7\textwidth]{eeg_nr_tsr.pdf} 
 %   \caption{Topography plots of mean EEG activity for both reading tasks, for a full sentence containing a \textit{founder} relation (top), and for the single decisive word of this relation type ``founded'' (bottom). The sentence read in this example is ``Henry Ford, with his son Edsel, founded the Ford Foundation in 1936 as a local philanthropic organization with a board charter to promote human welfare.''
 %   }
  %  \label{fig:topoplots}
%\end{figure*}

\begin{figure}[t]
    \centering
    \includegraphics[width=0.5\textwidth]{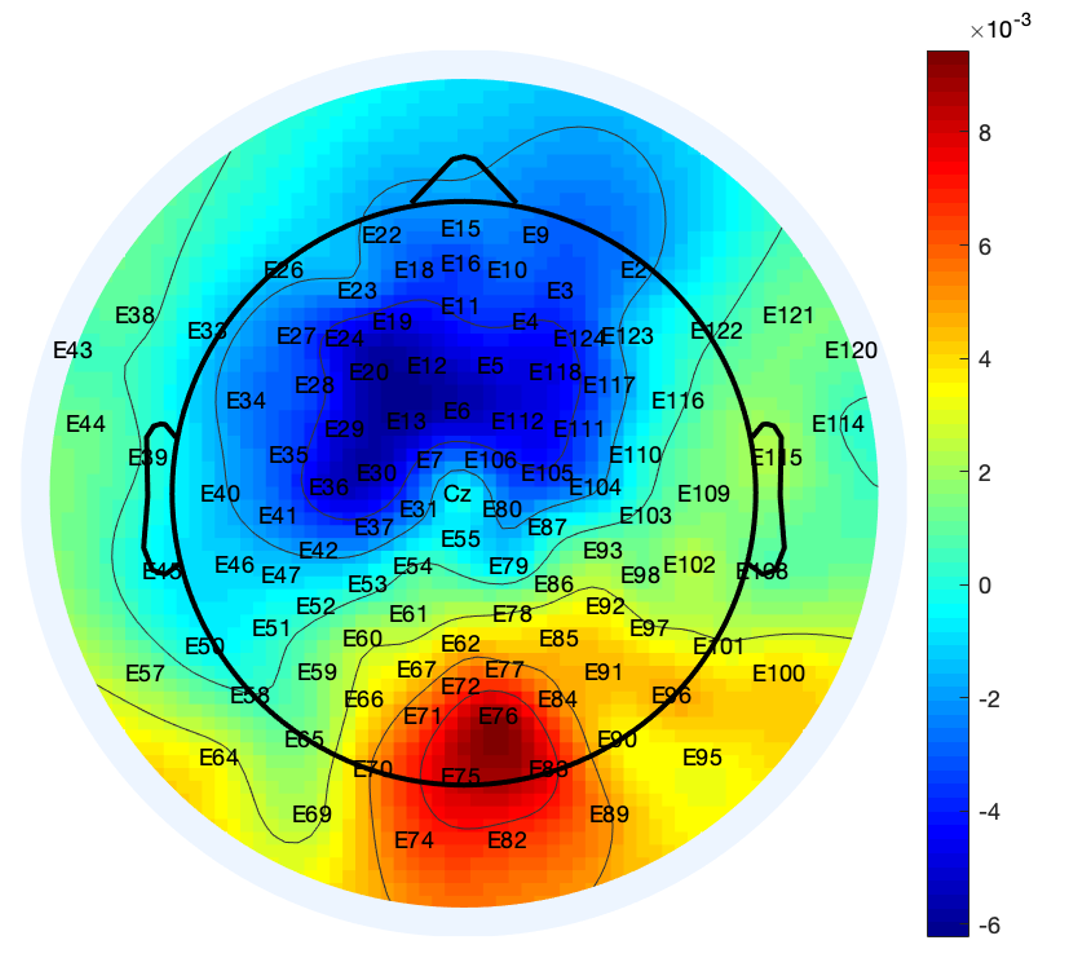} 
    \caption{Topography plot of the difference in the EEG activity between both reading tasks, averaged over all sentences all all subjects of the dataset (scalp viewed from above, nose at the top).
    }
    \label{fig:topoplots}
\end{figure}

\paragraph{EEG}
The EEG data shared in this project are available as raw data, but also preprocessed with Automagic (version 1.4.6, \newcite{pedroni2019automagic}), a tool for automatic EEG data cleaning and validation. 105 EEG channels (i.e. electrodes) were used from the scalp recordings. 9 EOG channels were used for artifact removal and additional 14 channels lying mainly on the neck and face were discarded before data analysis. Bad channels were identified and interpolated. We used the Multiple Artifact Rejection Algorithm (MARA), a supervised machine learning algorithm that evaluates ICA components, for automatic artifact rejection. MARA has been trained on manual component classifications, and thus captures a wide range of artifacts. MARA is especially effective at detecting and removing eye and muscle artifact components. The effect of this preprocessing can be seen in Figure \ref{fig:example}d.\\
After preprocessing, we synchronized the EEG and eye-tracking data to enable EEG analyses time-locked to the onsets of fixations. To compute oscillatory power measures, we band-pass filtered the continuous EEG signals across an entire reading task for five different frequency bands resulting in a time-series for each frequency band. The independent frequency bands were determined as follows: \textit{theta$_1$} (4–6 Hz), \textit{theta$_2$} (6.5–8 Hz), \textit{alpha$_1$} (8.5–10 Hz), \textit{alpha$_2$} (10.5–13 Hz), \textit{beta$_1$} (13.5–18 Hz), \textit{beta$_2$} (18.5–30 Hz), \textit{gamma$_1$} (30.5–40 Hz), and \textit{gamma$_2$} (40–49.5 Hz). We then applied a Hilbert transformation to each of these time-series. We specifically chose the Hilbert transformation to maintain the temporal information of the amplitude of the frequency bands, to enable the power of the different frequencies for time segments defined through the fixations from the eye-tracking recording. Thus, for each eye-tracking feature we computed the corresponding EEG feature in each frequency band. Furthermore, we extracted sentence-level EEG features by calculating the power in each frequency band, and additionally, the difference of the power spectra between frontal left and right homologue electrodes pairs. For each eye-tracking based EEG feature, all channels were subject to an artifact rejection criterion of $90\mu V$ to exclude trials with transient noise.

\section{Data Validation}\label{validation}

The aim of the technical validation of the data is to guarantee good recording quality and to replicate findings of previous studies investigating co-registration of EEG and eye movement data during natural reading tasks (e.g. \newcite{dimigen2011coregistration}). We also compare the results to ZuCo 1.0 \cite{hollenstein2018zuco}, which allows a more direct comparison due to the analogous recording procedure.

\begin{figure*}[t]
    \centering
    \includegraphics[width=1\textwidth]{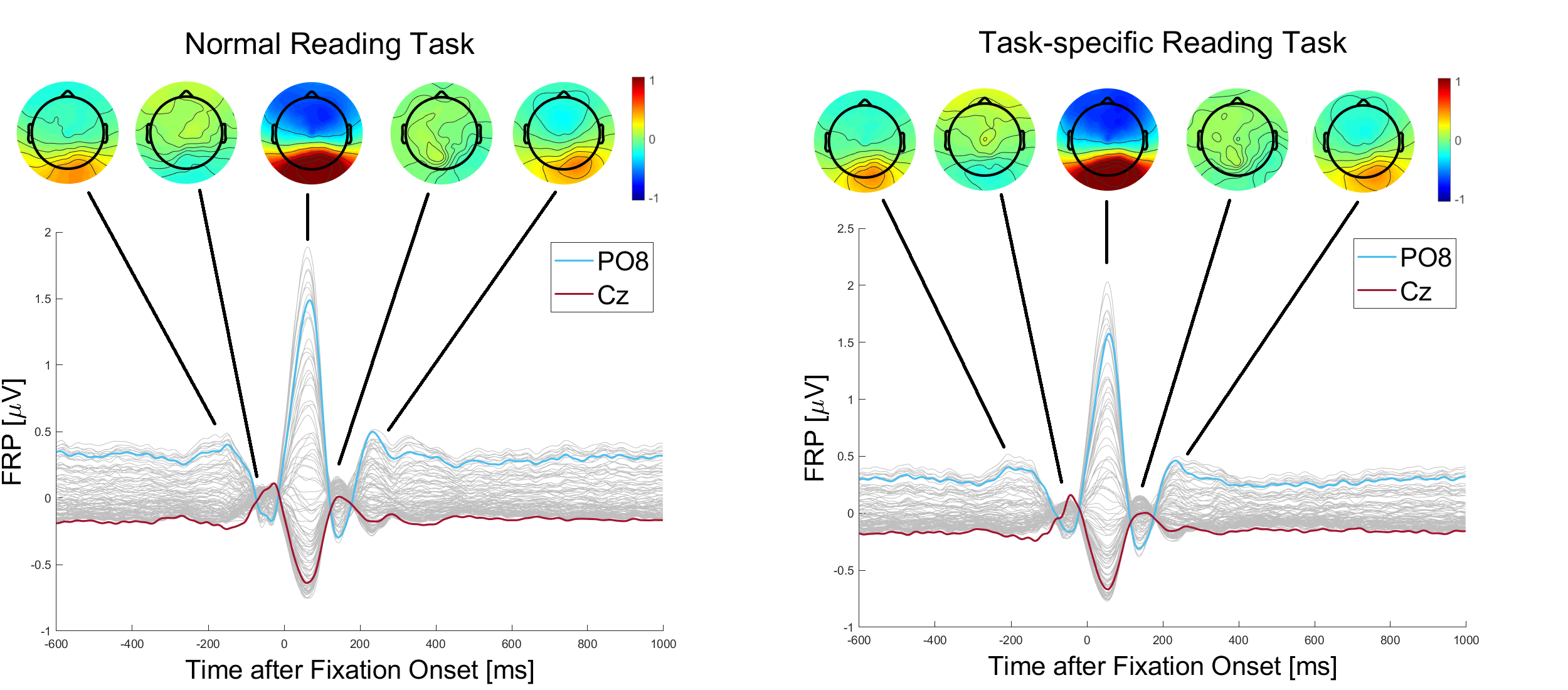} 
    \caption{Fixation-related potentials (FRPs) during both task conditions with selected scalp level potential distributions. Topographies show color-coded amplitudes in microvolt.}
    \label{fig:frps}
\end{figure*}

\begin{figure*}[t]
    \centering
    \includegraphics[width=1\textwidth]{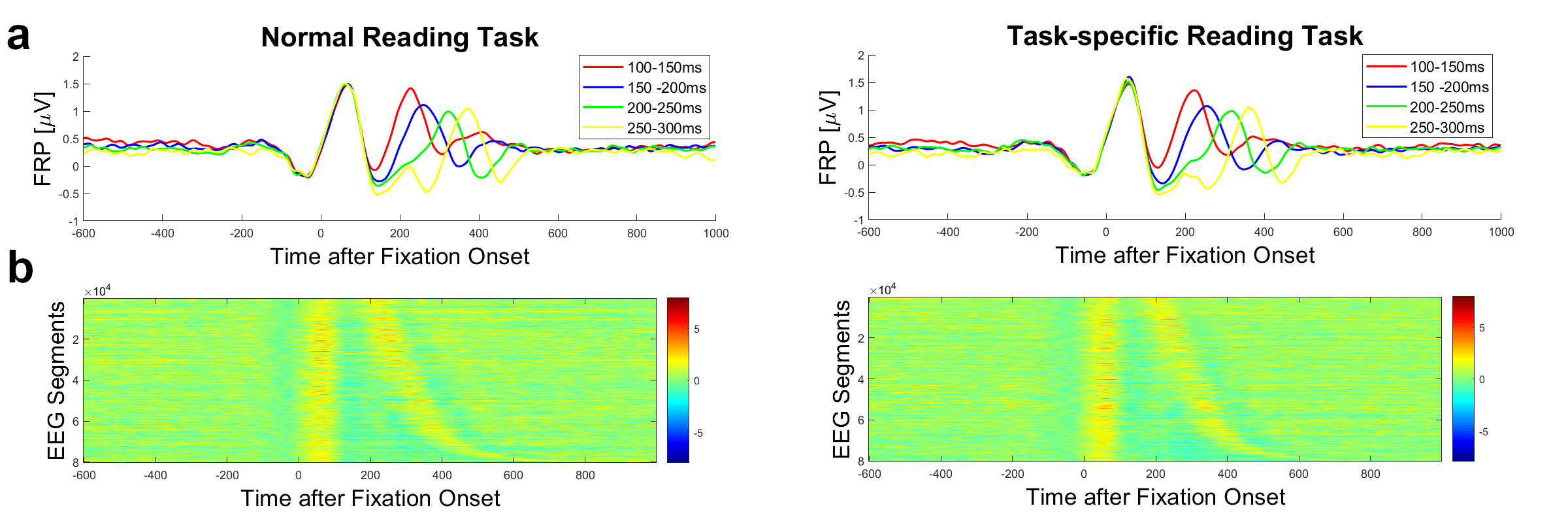} 
    \caption{Clustered EEG segments. (a) FRPs of electrode \textit{Cz}, clustered by duration of the fixation. (b) Each horizontal line represents the mean of the current and 50 adjacent EEG epochs, segmented on fixation onset. Segments are ordered by fixation duration (top: shortest fixation, bottom: longest fixation). Color represents the amplitude of the signal in microvolt.}
    \label{fig:fixsegs}
\end{figure*}

\paragraph{Eye-tracking}
We validated the recorded eye-tracking data by analyzing the fixations made by all subjects through their reading speed and omission rate on sentence level. The omission rate is defined as the percentage of words that is not fixated in a sentence. Figure \ref{fig:omr} (middle) shows the  mean reading speed over all subjects, measured in seconds per sentence and Figure \ref{fig:omr} (right) shows the mean omission rates aggregated over all subjects for each task. Clearly, the participants made less fixations during the task-specific reading, which lead to faster reading speed.

Moreover, we corroborated these sentence-level metrics by visualizing the skipping proportion on word level (Figure \ref{fig:skipping}). The skipping proportion is the average rate of words being skipped (i.e. not being fixated) in a sentence. As expected, this also increases in the task-specific reading.

Although the reading material is from the same source and of the same length range (see Figure \ref{fig:omr} (left)), in the first task (NR) passive reading was recorded, while in the second task (TSR) the subjects had to annotate a specific relation type in each sentence. Thus, the task-specific annotation reading lead to shorter passes because the goal was merely to recognize a relation in the text, but not necessarily to process the every word in each sentence. This distinct reading behavior is shown in Figure \ref{fig:heatmaps}, where fixations occur until the end of the sentence during normal reading, while during task-specific reading the fixations stop after the decisive words to detect a given relation type. Finally, we also analyzed the average reading times for each of the extracted eye-tracking features. The means and distributions for both tasks are shown in Figure \ref{fig:violinplots}. These results are in line with the recorded data in ZuCo 1.0, as well as with the features extracted in the GECO corpus \cite{cop2017presenting}.

\paragraph{EEG} The differences between normal reading and task-specific annotation reading is also evident in the brain activity data (see Figure \ref{fig:topoplots}).\\
As a first validation step, we extracted fixation-related potentials (FRPs), where the EEG signal during all fixations of one task are averaged. Figure \ref{fig:frps} shows the time-series of the resulting FRPs for two electrodes (\textit{PO8} and \textit{Cz}), as well as topographies of the voltage distributions across the scalp at selected points in time. The five components (for which the scalp topographies are plotted) are highly similar in the time-course of the chosen electrodes to \newcite{dimigen2011coregistration} as well as to ZuCo 1.0.

Moreover, these previous studies were able to show an effect of fixation duration on the resulting FRPs. To show this dependency we followed two approaches. First, for each reading task, all single-trial FRPs were ordered by fixation duration and a vertical sliding time-window was used to smooth the data \cite{dimigen2011coregistration}. Figure \ref{fig:fixsegs} (bottom) shows the resulting plots. In line with this previous work, a first positivation can be identified at 100 ms post-fixation onset. A second positive peak is located dependent on the duration of the fixation, which can be explained by the time-jittered succeeding fixation. The second approach is based on \newcite{henderson2013co} in which single trial EEG segments are clustered by the duration of the current fixation. As shown in Figure \ref{fig:fixsegs} (top), we chose four clusters and averaged the data within each cluster to four distinct FRPs, depending on the fixation duration. Again, the same positivation peaks become apparent. Both findings are consistent with the previous work mentioned and with our findings from ZuCo 1.0.

\section{Conclusion}

We presented a new, freely available corpus of eye movement and electrical brain activity recordings during natural reading as well as during annotation. This is the first dataset that allows for the comparison between these two reading paradigms.  We described the materials and experiment design in detail and conducted an extensive validation to ensure the quality of the recorded data. Since this corpus is tailored to cognitively-inspired NLP, the applications and re-use potentials of this data are extensive. The provided word-level \textit{and} sentence-level eye-tracking and EEG features can be used to improve and evaluate NLP and machine learning methods, for instance, to evaluate linguistic phenomena in neural models via neurolinguistic data. For instance, human language processing recordings can be used to probe the syntactic skills of language models \cite{toneva2019interpreting} or to evaluate the cognitive plausibility of word representations \cite{hollenstein2019cognival}. In addition, because the sentences contains semantic relation labels and the annotations of all participants, it can also be widely used for relation extraction and classification. Finally, the two carefully constructed reading paradigms allow for the comparison between normal reading and reading during annotation, which can be relevant to improve the manual labelling process as well as the quality of the annotations for supervised machine learning.

\section{Bibliographical References}
\bibliographystyle{lrec}
\bibliography{eeg-eyemove-nlp-oct2017}

%\section{Language Resource References}
%\label{lr:ref}
%\bibliographystylelanguageresource{lrec}
%\bibliographylanguageresource{languageresource}

\end{document}